\documentclass{article}
\usepackage{arxiv}
\usepackage[square,numbers]{natbib}
\usepackage[utf8]{inputenc} 
\usepackage[T1]{fontenc}    
\usepackage{hyperref}       
\usepackage{url}            
\usepackage{booktabs}       
\usepackage{amsfonts}       
\usepackage{nicefrac}       
\usepackage{microtype}      
\usepackage{lipsum}     
\usepackage{graphicx}
\usepackage{doi}
\usepackage{tabularx}
\usepackage{graphicx}
\usepackage{subcaption}
\usepackage{amsmath}
\usepackage{multirow}
\usepackage{longtable} 

\title{Towards Clinical Prediction with Transparency: An Explainable AI Approach to Survival Modelling in Residential Aged Care}

\author{ \href{https://orcid.org/0000-0001-9416-1435}{\includegraphics[scale=0.06]{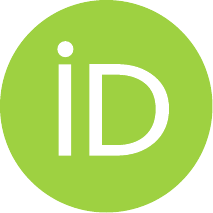}\hspace{1mm}Teo ~Susnjak} \\
    School of Mathematical and Computational Sciences\\
    Massey University\\
    Albany, New Zealand \\
    \texttt{t.susnjak@massey.ac.nz} \\
    \And
    {Elise~ Griffin} \\
    School of Mathematical and Computational Sciences\\
    Massey University\\
    Palmerston North, New Zealand \\
    \texttt{Elise.Griffin.1@uni.massey.ac.nz} \\
}

\renewcommand{\shorttitle}{Survival prediction in residential aged care patients using transparent machine learning}

\hypersetup{
pdftitle={A template for the arxiv style},
pdfsubject={q-bio.NC, q-bio.QM},
pdfauthor={David S.~Hippocampus, Elias D.~Striatum},
pdfkeywords={First keyword, Second keyword, More},
}

\begin{document}
\maketitle

\begin{abstract}
\textbf{Background:}  An accurate estimate of expected survival time assists people near the end of life to make informed decisions about their medical care.

\textbf{Objectives:} Use advanced machine learning methods to develop an interpretable survival model for older people admitted to residential age care.

\textbf{Setting:} A large Australasian provider of residential age care services.

\textbf{Participants:} All residents aged 65 years and older admitted for long-term residential care between July 2017 and August 2023.

\textbf{Sample size:}  11,944 residents from 40 individual care facilities.

\textbf{Predictors:} Age category, gender, and predictors related to falls, health status, co-morbidities, cognitive function, mood state, nutritional status, mobility, smoking history, sleep, skin integrity, and continence.

\textbf{Outcome:} Probability of survival at all time points post-admission.  The final model is calibrated to estimate the probability of survival at 6 months post-admission.

\textbf{Statistical Analysis:} Cox Proportional Hazards (CoxPH), Elastic Net (EN), Ridge Regression (RR), Lasso, Gradient Boosting (GB), XGBoost (XGB) and Random Forest (RF) were tested in 20 experiments using different train/test splits at a 90/10 ratio. Model accuracy was evaluated with the Concordance Index (C-index), Harrell's C-index, dynamic AUROC, Integrated Bier Score (IBS) and calibrated ROC analysis.  XGBoost was selected as the optimal model and calibrated for time-specific predictions at 1,3,6 and 12 months post admission using Platt scaling. SHapley Additive exPlanations (SHAP) values from the 6-month model were plotted to demonstrate the global and local effect of specific predictors on survival probabilities.

\textbf{Results:} For predicting survival across all time periods the GB, XGB and RF ensemble models had the best C-Index values of 0.714, 0.712 and 0.712 respectively.  We selected the XGB model for further development and calibration and to provide interpretable outputs.  The calibrated XGB model had a dynamic AUROC, when predicting survival at 6-months, of 0.746 (95\% CI 0.744-0.749). For individuals with a 0.2 survival probability  (80\% risk of death within 6-months) the model had a negative predictive value of 0.74. Increased age, male gender, reduced mobility, poor general health status, elevated pressure ulcer risk, and lack of appetite were identified as the strongest predictors of imminent mortality.

\textbf{Conclusions:} This study demonstrates the effective application of machine learning in developing a survival model for people admitted to residential aged care. The model has adequate predictive accuracy and confirms clinical intuition about specific mortality risk factors at both the cohort and the individual level.  Advancements in explainable AI, as demonstrated in this study, not only improve clinical usability of ML models by increasing transparency about how predictions are generated but may also reveal novel clinical insights.

\end{abstract}


\section{Introduction}
Predicting death is easy. Everybody will die. Estimating the precise probability of death for an individual within a specific time period is more difficult.   An accurate estimate of expected survival time helps people choose treatments that align with their goals of care \cite{weeks1998relationship}.  A falsely optimistic prognosis reduces the quality of death experienced by a patient and their loved ones \cite{christakis2000extent}.

Palliative care in people with a terminal diagnosis focuses on withdrawing treatments that cause pain or suffering and offering care that enhances the quality of remaining life. Many people are willing to endure short-term discomfort to increase survival time but there comes a point when sacrificing quality for quantity is no longer justifiable. For some, this realization comes just days before death, while for others, it may be recognized several months prior \cite{wright2008associations}.

People entering residential aged care do not usually have a specific terminal illness.  Rather, they are undergoing the inexorable decline in function that accompanies chronic illness and natural aging\cite{luppa2010prediction}. Shifting from an active treatment model to a palliative approach in this setting is a nuanced decision and is not always clearly communicated with residents or their families \cite{omori2022language}. However, more than one-third of older people admitted to residential aged care will die within six months of admission\cite{kelly2010length}. In most healthcare settings, this prognosis would prompt discussions about the patient’s care preferences in view of their short life expectancy. Yet these vital conversations occur less frequently than many older people would prefer \cite{sharp2013elderly}.

This study has two primary motivations. Firstly, we aim to employ advanced machine learning techniques to develop a reliable and accurate prognostic model for individuals entering residential aged care. By identifying those with a limited prognosis upon admission, we hope to empower healthcare providers to initiate transparent discussions with residents and their families about their end-of-life preferences. Secondly, our more expansive goal is to advocate for residential age care as a central provider of palliative services for those nearing the natural end of life. By emphasizing the limited life expectancy of people admitted to residential care and promoting open dialogue about it, we hope to enhance quality-of-care for all residents.

\section{Background}

Traditional methods in clinical decision support use observational epidemiological data and conventional statistical methods to create regression-based risk prediction models \cite{woodman2023comprehensive}. In general, these traditional models, while highly interpretable due to their use of model coefficients, are constrained by assumptions about data distributions, linearity, and a failure to include interactions among variables. They also depend heavily on domain expertise and usually utilise a small number of variables. In a recent review \citet{woodman2023comprehensive} report that these constraints result in generalized population-level models primarily intended to determine the predictive value of risk factors and the mean risk for individuals possessing a specific combination of these factors. Consequently, these models are less useful for personalized risk prediction and treatment recommendations, essential aspects of contemporary geriatric medicine \cite{martens2023personalized}.

Machine learning (ML) techniques have the potential to overcome the limitations of traditional approaches to survival analysis in geriatric populations   \cite{olender2023application}.  Recent studies have used ML algorithms to predict survival outcomes in people requiring palliative care, people with traumatic brain injury, and post-operative hip-fracture patients.  \cite{blanes2021responsive,wang2023prediction, Xing2022rf} \citet{blanes2021responsive} used Gradient Boosting with 20 variables from the patients' electronic health records to predict 1-year mortality in order to assess palliative care needs. \citet{Xing2022rf} explored Random Forest for predicting 1-year post-operative mortality risk for geriatric patients and concluded the model's efficacy.  \citet{wang2023prediction} compared the accuracies of Decision Tree, Random Forest, Support Vector Machine, Naïve Bayes, AdaBoost and XGboost to predict 30-day mortality for geriatric patients with a traumatic brain injury.

In studies involving cancer patients,  \citet{parikh2019predict} used ML algorithms to identify patients with cancer who are at risk of short-term mortality (6 months) in order to initiate timely conversations about treatment and end-of-life preferences. The study explored Gradient Boosting, Random Forest and Logistic Regression with clinicians concluding that the predictive models were effective. Similarly, \citet{manz2020ml} used Gradient Boosting to accurately predict 6-month mortality for patients with cancer using electronic health records, and was found to routinely outperform existing prognostic indices and was thus effective at identifying palliative care needs.

\citet{spooner2020comparison} compared ten ML algorithms for survival analysis in dementia prediction, utilizing high-dimensional clinical data from the Sydney Memory and Ageing Study (MAS) and the Alzheimer's Disease Neuroimaging Initiative (ADNI). These models demonstrated high concordance index values, signifying their accuracy in predicting dementia development. Similarly, \citet{wang2019development} developed a deep learning algorithm using longitudinal electronic health records from Partners HealthCare System to predict mortality risk in dementia patients. Their approach, analyzing various factors like palliative care and cognitive function, showed the potential of deep learning in mortality prediction. \citet{deardorff2022development} further validated a mortality prediction model for community-dwelling older adults with dementia, incorporating diverse predictors such as demographic and health factors, achieving good accuracy using the traditional Cox proportional hazards regression.

 The evolution from traditional regression-based models to advanced deep learning models will contribute to new clinical decision support tools capable of accommodating patient heterogeneity, including current and historic diagnoses, clinical events, medicine regimens, and measures of physical and cognitive function.  The current interest in ML for survival analysis is due to its efficacy in handling complex, multicollinear, high-dimensional data, a task where traditional methods falter.  ML algorithms, such as Random Forests and Deep Learning, excel in capturing complex interactions and non-linear relationships, leading to improved predictive accuracy, discovery of novel risk factors, and a deeper insight into underlying data relationships\cite{parikh2019predict,tran2022helicobacter,spooner2020comparison,wang2019development}.  Older people admitted to residential facilities and the clinicians who care for them may have much to gain by using decision-support tools developed through ML   \cite{woodman2023comprehensive}.

ML approaches in survival analysis, however, are not without limitations. They can be computationally demanding and possess "black box" characteristics, making their interpretability a significant challenge. This lack of transparency is a critical issue, especially in healthcare, where understanding the rationale behind predictions is essential for clinical interpretation. Addressing these challenges, SHAP (Shapley Additive exPlanations) and other interpretable AI techniques have emerged as solutions to enhance the transparency of ML models. \citet{wang2019development,blanes2021responsive} effectively used SHAP values to interpret the significance of various features in mortality risk predictions. Similarly, \citet{mostafaei2023machine}, employed SHAP values in their ML algorithms to identify key variables associated with mortality risk following a dementia diagnosis. These techniques provide insights into the contribution of each predictor to the model's predictions, enhancing the model's transparency and clinical utility for decision-making.

\subsubsection*{Objectives}
\begin{enumerate}
    \item Establish the feasibility of developing robust survival models using data acquired about patients during their first 31 days of admission to long-term care facilities.
    \item Determine the potential of various machine learning algorithms for survival predictions over various time horizons, aiming to identify the optimal algorithm in this context.
    \item Calibrate the predictive models to accurately forecast survival probabilities at a six-month time horizon post-admission, facilitating the optimization of targeted palliative care strategies.
    \item Demonstrate how explainable AI techniques increase the transparency and interpretability of predictive models, enhancing their utility in clinical decision-making.
    \item Propose an integrated framework combining predictive modelling, model interpretation, calibrated forecasts and clinical decision principles to optimize real-world application of the survival models.

\end{enumerate}

\subsubsection*{Contribution}
The contribution of this research is the development of a suite of predictive models that are methodologically robust and clinically actionable.  A key feature of our work is the integration of eXplainable AI (XAI) techniques which expose the internals of "black box" models generated by machine learning algorithms.  By quantifying the relative contribution of specific predictors to the prognosis of individual residents, these tools should increase confidence in the modelling among clinicians.  Generating a unique survival curve and identifying the salient risk factors for any individual admitted to residential age care is a novel feature of this work. An individualised survival curve could be used as a visual decision aid in clinical discussions about prognosis and patient preferences for end-of-life care.

\section{Methods}

This study is reported according to the TRIPOD (Transparent Reporting of a multivariable prediction model for Individual Prognosis Or Diagnosis) guideline \cite{collins2015transparent}.

\subsection{Data Source}
Health care data used in this study was  collected during the provision of routine care to older individuals admitted for long-term care between 1st July 2017 and 30th August 2023 to facilities owned by a single large Australasian private residential aged care provider.

\subsection{Participants}
Data from residents at 34 New Zealand and 6 Australian aged care facilities were included. Individuals were eligible for inclusion if they were admitted for long-term care on or after 1st July 2017.

\subsection{Outcome}
The primary outcome is the survival probability of an individual resident at admission to a residential age care facility assessed in two ways.

\begin{enumerate}
    \item A continuous survival curve showing survival probability at all time points up to six years post-admission
    \item A point-estimate of the probability of survival at six-months post-admission

\end{enumerate}

\subsection{Predictors}
All predictors included in the model were selected from demographic and clinical data recorded by registered nurses during the initial clinical evaluation of newly admitted residents.  Medication data was taken from the electronic medicine chart.   The earliest instance of each specific predictor following admission was used.  Any predictor recorded more than 31 days after admission was excluded. Our objective was to predict survival probability from the time of admission.  Data collected more than one month post-admission was discarded.

Tables \ref{tab:cohort_demographics} and \ref{tab:predictors} show the list of all predictors included in the model and the values assigned to each level. Table \ref{tab:cohort_demographics} details the demographic attributes of the study cohort, including reasons for discharge and Rx-Risk Co-morbidity Index diagnostic categories \cite{pratt2018validity}. Table \ref{tab:predictors} reports clinical variables drawn from the initial nursing assessment. Domain expertise was used to assign ordinal values representing the degree of severity for each level of the predictor.   High values represent a state associated with worse function or clinical status (higher mortality risk) and low values represent states associated with better function or clinical status (lower mortality risk).

\begingroup
\small
\begin{longtable}[c]{@{}rrrc@{}}
\caption{Cohort Demographics}
\label{tab:cohort_demographics}\\
\toprule
\textbf{Category} & \textbf{Residents (n)} & \textbf{Residents (\%)} & \textbf{Value} \\* \midrule
\endfirsthead
\multicolumn{4}{c}%
{{\bfseries Table \thetable\ continued from previous page}} \\
\endhead
\bottomrule
\endfoot
\endlastfoot

\addlinespace
\multicolumn{4}{l}{\textit{Age (years)}} \\
65-69 & 248 & 2\% & 67 \\
70-74 & 645 & 5\% & 72 \\
75-79 & 1435 & 12\% & 77 \\
80-84 & 2445 & 20\% & 82 \\
85-89 & 3255 & 27\% & 87 \\
90-94 & 2725 & 23\% & 92 \\
95-99 & 1060 & 9\% & 97 \\
100+ & 130 & 1\% & 100 \\
Age (mean,SD) & 85.7 (mean) & 7.2 (SD) & \\
\cmidrule(l){1-4}
\addlinespace
\multicolumn{4}{l}{\textit{Gender}} \\
Female & 7200 & 60\% & 0 \\
Male & 4494 & 38\% & 1 \\
Other/Gender Diverse & 167 & 1\% & 0 \\
Unknown & 82 & 1\% & 0 \\
\cmidrule(l){1-4}
\addlinespace
\multicolumn{4}{l}{\textit{Discharge Reason}} \\
Deceased & 6725 & 56\% & 1 \\
Current resident & 3465 & 29\% & 0 \\
Transfer to another care facility & 1145 & 10\% & 0 \\
Discharged home & 351 & 3\% & 0 \\
Transfer to public hospital & 244 & 2\% & 0 \\
Transfer to hospice & 14 & <1\% & 1 \\
\cmidrule(l){1-4}
\addlinespace
\multicolumn{4}{l}{\textit {Rx-Risk Comorbidity Index}\cite{pratt2018validity}}\\
Pain & 7151 & 79\% & 3 \\
Psychotic disorder & 3080 & 34\% & 6 \\
Congestive heart failure & 2383 & 26\% & 2 \\
Gastrooesophageal reflux disease & 2213 & 24\% & 0 \\
Ischemic heart disease:hypertension & 1751 & 19\% & -1 \\
Depression & 1693 & 19\% & 2 \\
Antiplatelets & 1449 & 16\% & 2 \\
Anticoagulants & 1132 & 12\% & 1 \\
Hyperlipidaemia & 1030 & 11\% & -1 \\
Anxiety & 958 & 11\% & 1 \\
Chronic airways disease & 932 & 10\% & 2 \\
Allergies & 772 & 9\% & -1 \\
Steroid-responsive disease & 757 & 8\% & 2 \\
Diabetes & 636 & 7\% & 2 \\
Ischemic heart disease & 581 & 6\% & 2 \\
Hypertension & 540 & 6\% & -1 \\
Dementia & 498 & 5\% & 2 \\
Glaucoma & 469 & 5\% & 0 \\
Hypothyroidism & 430 & 5\% & 0 \\
Gout & 403 & 4\% & 1 \\
Arrhythmia & 325 & 4\% & 2 \\
Malignancies & 325 & 4\% & 2 \\
Osteoporosis/Pagets & 321 & 4\% & -1 \\
Inflammation/pain & 308 & 3\% & -1 \\
Parkinsons disease & 306 & 3\% & 3 \\
Benign prostatic hypertrophy & 279 & 3\% & 0 \\
Incontinence & 247 & 3\% & 0 \\
Epilepsy & 246 & 3\% & 0 \\
Benign prostatic hyperplasia & 200 & 2\% & 0 \\
Renal disease & 59 & 1\% & 6 \\
Smoking cessation & 59 & 1\% & 6 \\
Autoimmune and rheumatological conditions & <50 & <1\% & 0 \\
Hyperthyroidism & <50 & <1\% & 2 \\
Psoriasis & <50 & <1\% & 0 \\
Inflammatory bowel disease & <50 & <1\% & 0 \\
Malnutrition & <50 & <1\% & 0 \\
Bipolar disorder & <50 & <1\% & -1 \\
Pancreatic insufficiency & <50 & <1\% & 0 \\
Migraine & <50 & <1\% & -1 \\
Liver failure & <50 & <1\% & 3 \\
Transplant & <50 & <1\% & 0 \\
Hepatitis B & <50 & <1\% & 0 \\
Hyperkalemia & <50 & <1\% & 4 \\
Pulmonary hypertension & <50 & <1\% & 6 \\
\bottomrule
\end{longtable}
\endgroup

Predictors relating to nutrition, mobility, smoking, sleep, skin integrity and continence are from a standardised set of questions and answers based on the InterRAI long-term care facilities assessment \cite{interRAI}. Predictors relating to current health status, cognition, mood and pressure ulcers are drawn from validated InterRAI-based composite measures (respectively, the CHESS scale (Changes in Health, End-Stage Disease, Symptoms and Signs) \cite{hirdes2014use, ogarek2018minimum}, cognitive performance scale \cite{travers2013validation}, depression rating scale \cite{penny2016convergent}, pressure ulcer risk scale \cite{poss2010development}).  Co-morbidities are assessed in two ways. The presence of a specific diagnosis was established by filtering the diagnosis fields in the resident clinical record for terms that captured dementia of any type, ischemic cardiac disease and heart failure of any type, any malignant neoplasm, any non-cancer pulmonary disease and any form of diabetes, excluding glucose intolerance. The sum of the scores for these items (1 = ANY diagnosis of this type present, 0 = ALL diagnoses of this type absent), rather than the individual diagnosis, was used as a predictor in the final model (minimum value = 0, maximum value =5).  The Rx-Risk Co-morbidity Index is included as a second comorbidity item. This index provides a weighted score for each specific diagnosis based on prescription data.  The scores are summed for an individual resident to provide the final Rx-Risk score. The falls predictor was a bespoke question used by the provider about the frequency of falls in the past six months.

\subsection{Sample Size}
Sample size was determined by the availability of data.  Complete digital personal health records for all residents, including electronic medicine chart data, were available from July 1st 2017.  We utilised all data from current and discharged long-term residents admitted on or after this date for model development.

\subsection{ Missing Data}
Missing data was encountered at varying degrees for most predictors, as reported in Table \ref{tab:predictors}.  Predictors with 75\% or more missing values were excluded. The Multiple Imputation by Chained Equations (MICE) \cite{raghunathan2001multivariate,van2007multiple} was used to impute missing values for all predictors, motivated by recent studies demonstrating this approach in the context of survival analyses \cite{qi2018strategies,mera2021evaluating}. Table \ref{tab:predictors} reports the percentage of missing values for each included predictor.

\begin{table}[htbp]
\centering
\scriptsize
\caption{Predictor Variables and Values}
\label{tab:predictors}

\setlength{\extrarowheight}{-0.2pt} 
\setlength{\aboverulesep}{1pt} 
\setlength{\belowrulesep}{1pt} 

\begin{tabularx}{\textwidth}{@{}lXlrrc@{}}
\toprule
\textbf{Predictors} & \textbf{Assessment Question} & \textbf{Answer} & \textbf{Residents (n)} & \textbf{Residents (\%)} & \textbf{Value} \\
\midrule
\multirow{5}{*}{Falls} & \multirow{5}{=}{History of falls} & No history of falls & 5125 & 42.9\% & 0 \\
 & & 4 or less in last 6 months & 5270 & 44.1\% & 1 \\
 & & 5 or more in last 6 months & 477 & 4.0\% & 2 \\
 & & 3 or more falls in one month period & 306 & 2.6\% & 3 \\
 & & Missing & 766 & 6.4\% &  \\
\midrule
\multirow{7}{*}{Health status} & \multirow{7}{=}{What was the CHESS scale score?} & No symptoms & 1446 & 12.1\% & 0 \\
 & & Minimal health instability & 1512 & 12.7\% & 1 \\
 & & Low health instability & 1444 & 12.1\% & 2 \\
 & & Moderate health instability & 793 & 6.6\% & 3 \\
 & & High health instability & 385 & 3.2\% & 4 \\
 & & Highest level of instability & 65 & 0.5\% & 5 \\
 & & Missing & 6299 & 52.7\% &  \\
\midrule
\multirow{5}{*}{Comorbidities} & \multirow{2}{=}{What was the weighted Rx-Risk scale score?} & Sum of weighted scores (range -3 to 23) & 9065 & 75.9\% &  \\
 & & Missing & 2879 & 24.1\% &  \\
\cmidrule(l){2-6}
 & \multirow{5}{=}{Was this diagnosis present?} & Dementia & 5179 & 43.4\% & 1 \\
 & & Heart disease & 3658 & 30.6\% & 1 \\
 & & Cancer & 1301 & 10.9\% & 1 \\
 & & Diabetes & 1289 & 10.8\% & 1 \\
 & & Lung disease & 1112 & 9.3\% & 1 \\
\midrule
\multirow{4}{*}{Cognition} & \multirow{4}{=}{What was the cognitive performance scale score?} & Intact & 639 & 5.3\% & 0 \\
 & & Borderline intact & 538 & 4.5\% & 1 \\
 & & Mild impairment & 2127 & 17.8\% & 2 \\
 & & Moderate impairment & 1527 & 12.8\% & 3 \\
 & & Moderate/Severe impairment & 199 & 1.7\% & 4 \\
 & & Severe impairment & 426 & 3.6\% & 5 \\
 & & Very severe impairment & 99 & 0.8\% & 6 \\
 & & Missing & 6389 & 53.5\% &  \\
\midrule
\multirow{4}{*}{Mood} & \multirow{4}{=}{What was the depression rating scale score?} & None (0) & 2609 & 21.8\% & 0 \\
 & & Mild (1-2) & 1729 & 14.5\% & 1 \\
 & & Moderate (3-5) & 909 & 7.6\% & 2 \\
 & & Severe (6-14) & 308 & 2.6\% & 3 \\
 & & Missing & 6389 & 53.5\% &  \\
\midrule
\multirow{3}{*}{Nutrition} & \multirow{3}{=}{Has the resident lost weight recently?} & No & 4379 & 36.7\% & 0 \\
 & & Unsure & 5242 & 43.9\% & 0 \\
 & & Yes & 1650 & 13.8\% & 1 \\
 & & Missing & 673 & 5.6\% &  \\
\cmidrule(l){2-6}
 & \multirow{3}{=}{Is the resident eating poorly or has a lack of appetite?} & No & 9136 & 76.5\% & 0 \\
 & & Yes & 2135 & 17.9\% & 1 \\
 & & Missing & 673 & 5.6\% &  \\
\midrule
\multirow{7}{*}{Mobility} & \multirow{7}{=}{How does your resident mobilise?} & Independent & 4175 & 34.9\% & 0 \\
  & & Supervision or prompting & 1889 & 15.8\% & 1 \\
 & & 1 person assistance & 2395 & 20.1\% & 2 \\
 & & 2 person assistance & 921 & 7.7\% & 3 \\
 & & Does not mobilise (bed or chair bound) & 1153 & 9.7\% & 4 \\
 & & Missing & 1411 & 11.8\% &  \\
\cmidrule(l){2-6}
 & \multirow{7}{=}{What equipment does your resident use to mobilise safely?} & None & 2698 & 22.6\% & 0 \\
 & & Walking stick & 813 & 6.8\% & 1 \\
 & & Walking frame & 5083 & 42.6\% & 2 \\
 & & Transfer belt or other &  586 & 1.7\% & 3 \\
 & & Gutter frame & 220 & 1.8\% & 4 \\
 & & Wheelchair, fallout chair or lazyboy & 1130 & 9.5\% & 5 \\
 & & Missing & 1411 & 11.8\% &  \\
\midrule
\multirow{2}{*}{Smoking} & \multirow{2}{=}{Has your resident smoked in the past?} & No & 7604 & 63.7\% & 0 \\
 & & Yes & 1950 & 16.3\% & 1 \\
 & & Missing & 2390 & 20.0\% &  \\
\midrule
\multirow{2}{*}{Sleep} & \multirow{2}{=}{Does your resident require assistance to settle to bed at night?} & No & 3861 & 32.3\% & 0 \\
 & & Yes & 6631 & 55.5\% & 1 \\
 & & Missing & 1452 & 12.2\% &  \\
\midrule
\multirow{4}{*}{Skin} & \multirow{4}{=}{Has your resident's skin integrity changed since last assessment?} & Improved & 170 & 1.4\% & 0 \\
 & & No Change & 2609 & 21.8\% & 0 \\
 & & Fluctuated & 157 & 1.3\% & 1 \\
 & & Declined & 530 & 4.4\% & 2 \\
 & & Missing & 8478 & 71.0\% &  \\
\cmidrule(l){2-6}
 & \multirow{5}{=}{What was the Pressure Ulcer Risk scale?} & Very low risk & 2629 & 22.0\% & 0 \\
 & & Low risk & 1967 & 16.5\% & 1 \\
 & & Moderate risk & 554 & 4.6\% & 2 \\
 & & High risk & 349 & 2.9\% & 3 \\
 & & Very high risk & 34 & 0.3\% & 4 \\
 & & Missing & 6411 & 53.7\% &  \\
\midrule
\multirow{2}{*}{Continence} & \multirow{2}{=}{Is the resident incontinent of faeces?} & No & 3093 & 25.9\% & 0 \\
 & & Yes & 2187 & 18.3\% & 1 \\
 & & Missing & 6664 & 55.8\% &  \\
\cmidrule(l){2-6}
 & \multirow{2}{=}{Is the resident incontinent of urine?} & No & 3808 & 31.9\% & 0 \\
 & & Yes & 1554 & 13.0\% & 1 \\
 & & Missing & 6582 & 55.1\% &  \\
\bottomrule
\end{tabularx}
\end{table}

\subsection{Statistical Analysis Methods}

During the preliminary stages of our data processing, a pairwise correlation coefficient threshold of 0.7 was used as a guide for eliminating highly correlated variables to ensure model parsimony and reduce multicollinearity \cite{Shah2020clinical}.  The decision on which one of the variables to exclude from the model was made by examining data quality and completeness, the relative univariate predictive power of the variable, and potential interpretability. We standardized the data prior to use for the Lasso, Ridge, and Elastic Net algorithms.
The algorithms\footnote{The implementations of the algorithms from the Python library scikit-survival \cite{sksurv} version 0.21.0 were used and XGBoost \cite{xgboost176} version 1.7.6.} used for candidate models are described in Table \ref{tab:alg}.

\begin{table}[htb]
    \centering
    \caption{Overview of Machine Learning Models Employed in the Study}
    \begin{tabularx}{\textwidth}{lX}
    \toprule
    Model & Characteristics \\
    \midrule
    Cox Proportional Hazards \cite{tibshirani1997} & The CoxPH is an extension of the classical Cox Proportional Hazards model, containing a penalty term to better manage high-dimensional datasets. Its core advantage is its capacity to manage the problems of  both high dimensionality and event sparsity.\\
    \addlinespace
    Elastic Net \cite{zou2005} & The Elastic Net model amalgamates the L1 and L2 regularization techniques of Lasso and Ridge Regression, respectively. This hybridization allows the model to efficiently navigate the challenges of multicollinearity and variable selection . \\
    \addlinespace
    Ridge Regression \cite{hoerl1970} & Ridge Regression employs L2 regularization to provide an alternative approach to handling multicollinearity. It is adept at shrinking coefficients,  stabilizing them in the presence of highly correlated variables. \\
    \addlinespace
    Lasso \cite{tibshirani1996} & Lasso utilizes L1 regularization to achieve both regularization and variable selection. It is especially useful for high-dimensional datasets where feature selection is vital, as it drives some coefficients to zero. \\
    \addlinespace
    Gradient Boosting \cite{friedman2001} & Gradient Boosting is a state-of-the-art ensemble learning technique that builds strong predictive models by aggregating weak learners. Its adaptability and effectiveness have been empirically validated in many settings, including healthcare. \\
    \addlinespace
    XGBoost \cite{chen2016} & XGBoost is as an optimized variant of the Gradient Boosting algorithm, notable for computational efficiency and scalability. Its predictive capability has been demonstrated through its dominance in various ML competitions. \\
    \addlinespace
    Random Forest \cite{breiman2001} & Random Forest is an ensemble of decision trees, each constructed with a bootstrapped sample of the data and a subset of variables. Its robustness against outliers and irrelevant features makes it well suited for modelling clinical data. \\
    \bottomrule
    \end{tabularx}
    \label{tab:alg}
\end{table}

Before running experiments with these algorithms, hyperparameter tuning was executed using train/test splits. This process involved exploring a range of hyperparameters, guided by the goal of maximizing model performance. Table \ref{tab:alg_hyperparameters} summarizes the resulting hyperparameters used in the subsequent experiments for each algorithm, reflecting the settings that yielded the best results in our analyses.

\begin{table}[htb]
    \centering
    \caption{Overview of hyperparameter settings for the models}
    \begin{tabularx}{\textwidth}{lX}
    \toprule
    Model & Hyperparameters\\
    \midrule
    CoxPH \cite{tibshirani1997} & None \\
    \addlinespace
    Elastic Net  & l1\_ratio=1.0, n\_alphas=1, alphas=[0.00034], normalize=True, fit\_baseline\_model=True \\
    \addlinespace
    Ridge Regression  & l1\_ratio=$10^{-100}$, n\_alphas=1, alphas=[2.24e-06], normalize=True, fit\_baseline\_model=True \\
    \addlinespace
    Lasso  & l1\_ratio=0.9, alpha\_min\_ratio=0.01, fit\_baseline\_model=True \\
    \addlinespace
    Gradient Boosting  & n\_estimators=771, min\_samples\_split=20.04, max\_depth=7, min\_samples\_leaf=1.85, learning\_rate=0.28, dropout\_rate=0.05, objective='survival:cox', max\_features=4, subsample=0.83 \\
    \addlinespace
    XGBoost  & num\_boost\_round=1107, learning\_rate=0.018, max\_depth=3, colsample\_bytree=0.83, gamma=0.49, objective='survival:cox', subsample=0.58 \\
    \addlinespace
    Random Forest  & n\_estimators=592, min\_samples\_split=2.54, max\_depth=7, min\_samples\_leaf=20.89 \\
    \bottomrule
    \end{tabularx}
    \label{tab:alg_hyperparameters}
\end{table}

Subsequent to hyperparameter tuning, each algorithm was tested in 20 experiments using different train/test splits at a 90/10 ratio. For algorithms requiring a validation set, the training set was further divided using a 90/10 split. Outcomes from these experiments were aggregated and presented with 95\% confidence intervals.
Model performance was assessed using various evaluation measures, detailed in Table \ref{tab:eval}. The multi-metric approach enables the evaluation of both uncalibrated and time-specific metrics and offers a comprehensive understanding of each model's predictive accuracy, discriminative power, and reliability. These metrics serve both as individual model performance indicators and as tools for model comparison. The best-performing model was calibrated for time-specific predictions at a six-month time point using Platt scaling \cite{niculescu2005predicting,rajaraman2022calibration}.
The effectiveness of the calibration also entailed performing 20 train/test splits at a 90/10 ratio to evaluate the accuracy, which was visualized via a calibration plot and reported through Dynamic AUROC, IBS C-index and Harrell C-index.
The specificity, sensitivity and negative predictive power (accuracy in forecasting mortality for the people who died within six months of admission) of this model were inspected via ROC curve analyses.

\begin{table}[h!]
    \centering
    \caption{Overview of Evaluation Metrics Used in the Study}
    \begin{tabularx}{\textwidth}{lX}
    \toprule
    Metric & Description \\
    \midrule
    Concordance Index (C-index) \cite{harrell1982} & The Concordance Index (C-index) is a global metric for assessing the discriminative ability of survival models across the entire range of observed times. It quantifies the model's capability to correctly rank pairs of subjects based on their survival times, making it useful for a general assessment of risk over a period. \\
    \addlinespace
    Harrell's C-index \cite{harrell2015} & An extension of the standard C-index, Harrell's C-index accommodates the complexities of censored data and offers a robust evaluation metric across different risk strata. It is also a more robust metric to high rates of censoring, present in this study's dataset. \\
    \addlinespace
    Dynamic AUROC \cite{hanley1982} & Unlike the standard Area Under the Receiver Operating Characteristic Curve (AUROC), the Dynamic AUROC is time-specific and evaluates the model's discriminative ability at predetermined time points. It offers critical insights into the model's performance in separating those who will experience the event from those who won't at each specific time point. \\
    \addlinespace
    Integrated Brier Score (IBS) \cite{graf1999} & The Integrated Brier Score (IBS) is a time-specific metric that quantifies both the calibration and discrimination of the model. It provides an average measure of prediction error for survival probabilities at specific time points, thus offering a nuanced assessment of the model's predictive reliability. \\
    \addlinespace
    Calibrated ROC Analysis & The ROC curve analysis was performed on the calibrated model. The ROC curve was used to investigate the specificity and sensitivity of this model and to demonstrate the variation in predictive power at different threshold values for the probability of survival.   \cite{manz2020ml}. \\
    \bottomrule
    \end{tabularx}
    \label{tab:eval}
\end{table}

The final best-performing model from the analyses, its calibration together with an example code on how to use it, is publicly available from a GitHub repository\footnote{https://github.com/teosusnjak/survival-analysis-stage1}.

\subsection{Ethical Considerations}
Ethics approval for this study was granted by the Aotearoa Research Ethics Committee  (formerly New Zealand Ethics Committee, NZEC22\_11) and noted by the Human Ethics Committee  (Ohu Matatika 2) of Massey University.

\subsection{eXplainable AI Tools}

Balancing predictive strength and model interpretability is a specific challenge in contemporary machine learning research. As algorithms increase in complexity, the transparency and explainability of derived models diminish, creating potential concerns in settings, such as healthcare, where decisions based on model outputs must align with ethical and regulatory standards. eXplainable Artificial Intelligence (XAI) is an important line of research that addresses transparency in ML and comprises a suite of tools designed to expose the internal decision-making processes of advanced models. Of these tools, SHapley Additive exPlanations (SHAP), used in this study, is an exemplar of XAI methods \cite{gramegna2021shap, Lundberg2017}.

We report model behaviour from both \textit{global} (cohort-level) and \textit{local} (patient-level) interpretative dimensions. At a cohort-level, we provide SHAP Summary plots to show the ranked impact of predictors on survival probability and SHAP Dependence plots to illustrate the effect of interactions between predictors on predicted survival. These plots provide a \textit{macroscopic} lens into the primary determinants of the model predictions. For patient-level analyses, we use SHAP Waterfall plots. These plots provide a granular examination of individual data points, detailing the contribution of each predictor to a specific prognosis.  These plots provide patient-level information on the most relevant predictors in each individual person, making the algorithms' predictive processes transparent and increasing clinician confidence in the model output.

\section{Results}
\subsection{Overview}
We present our results in two parts. In the first section, we report performance metrics for a variety of uncalibrated general models trained to predict survival probability at all time points up to six years post-admission. We use individualised survival curves and SHAP summary, dependence and waterfall plots to provide  insight into the behaviour of the best-performing uncalibrated model. In the second part, we report performance metrics for a similar model, calibrated to predict survival probability at six months post-admission time.  Finally, we present the clinical evaluation metrics of sensitivity, specificity, and negative predictive value for the calibrated model.

\subsection{Participants}
Data from 12882 individuals were extracted from the database. We eliminated 407 individuals who lacked requisite assessment data within 31 days of admission from the cohort.  We reconciled data from residents with one or more consecutive admissions, resulting in a cohort of 11945 unique individuals. Data from one resident was excluded due to a negative value length of stay value . The final cohort contained data from 11944 individuals.   The mean age of people in the cohort was 86 years (SD 7) and the majority were women (n= 7200, 60\%).  Just over half the cohort (n = 6739, 56\%) were discharged due to death (the modelled outcome ) and approximately 30\% were current residents (n = 3465).  Three-quarters of residents had an electronic medicine chart initiated within 31 days of admission, allowing us to estimate the Rx-Risk Comorbidity Index for these individuals (n = 9065, 76\%) \cite{pratt2018validity}.  The most common diagnostic categories based on prescription data were pain, a psychotic disorder (most likely behavioural and psychological symptoms of dementia), congestive heart failure and gastro-oesophageal reflux.  The full distribution of Rx-Risk Comorbidity Index categories is reported in Table \ref{tab:cohort_demographics}.

\subsection{Model Performance}

Table \ref{tab:results} shows evaluation results of survival models across a time horizon of up to 74 months. The best-performing models, according to the C-index, are the ensemble methods, Gradient Boosting, Random Forest and XGBoost with negligible differences between them. These models exhibit C-indices of 0.712 to 0.714, supported by narrow 95\% confidence intervals, indicating effective discriminatory power and robust statistical stability. The leading C-index of the top three models is complemented by their Harrell's C-index score of $\sim$0.67 and an AUROC of $\sim$0.75, confirming effective performance in both discrimination and calibration. Only marginally lower, CoxPH, Ridge and Lasso regression exhibit similar performance on this dataset across all the metrics.  Elastic Net performed significantly worse than all other candidate models.

\begin{table}[hbt]
    \caption{Rank-ordered performance metrics of different models across the entire survival period up to 74 months post admission to a care facility showing 95\% confidence intervals across the key metrics.}
    \label{tab:results}
    \centering
    \begin{tabularx}{0.7\textwidth}{rXXX}
    \toprule
    Model & C-index & Harrell's C-index & AUROC \\
    \midrule
    GB & 0.714 (0.711-0.717) & 0.673 (0.67-0.676) & 0.747 (0.743-0.751) \\
    RF & 0.712 (0.708-0.715) & 0.671 (0.667-0.674) & 0.745 (0.739-0.75) \\
    XGB & 0.712 (0.709-0.716) & 0.675 (0.672-0.679) & 0.755 (0.75-0.759) \\
    CoxPH & 0.709 (0.706-0.711) & 0.671 (0.668-0.674) & 0.749 (0.746-0.752) \\
    Ridge & 0.708 (0.705-0.711) & 0.670 (0.667-0.673) & 0.745 (0.742-0.749) \\
    Lasso & 0.706 (0.704-0.709) & 0.666 (0.663-0.669) & 0.742 (0.739-0.746) \\
    Elastic & 0.532 (0.529-0.535) & 0.543 (0.54-0.546) & 0.555 (0.551-0.56) \\
    \bottomrule
    \end{tabularx}
\end{table}

\subsubsection*{Model Interpretability}

\begin{figure}[hbt]
    \centering
    \includegraphics[width=0.74\textwidth]{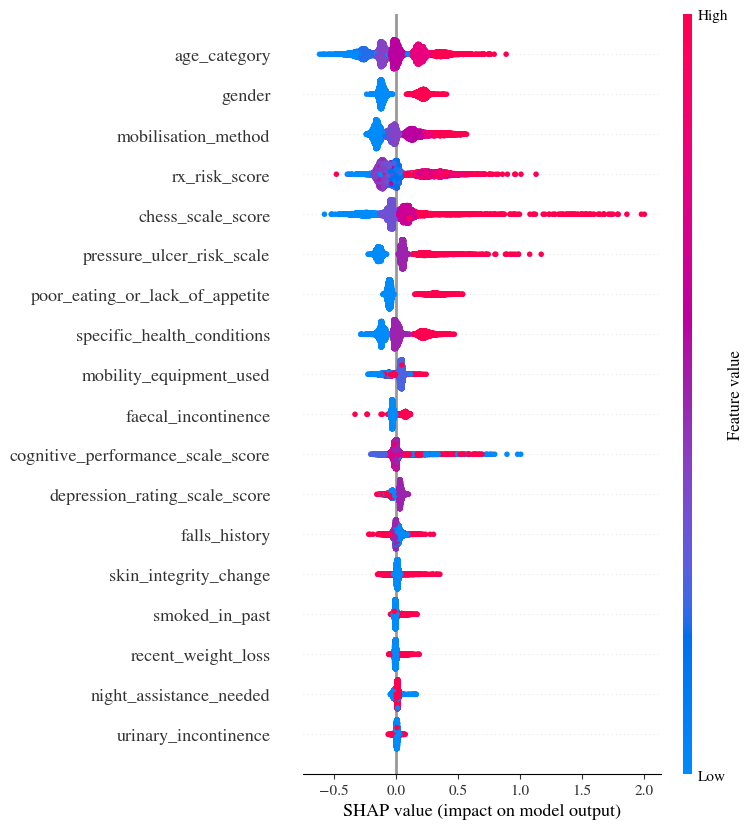}
    \caption{Predictor importance summary plot for the XGBoost model.}
    \label{fig:importance}
\end{figure}

Here we examine the internal mechanics of the model and the impact of each predictor at the cohort level and at the individual patient level. For this analysis, we selected XGBoost\footnote{While Gradient Boosting was technically the top performing model, both it and XGBoost are essentially the same algorithm with slightly different implementations. The implementation of XGBoost, however, lends itself better for interpretability analysis given its integration with SHAP tools. }, one of the top performing models according to Table \ref{tab:results},  for a more detailed inspection. The SHAP summary plot in Figure \ref{fig:importance} presents the predictors included in the model ranked in order of importance (greatest influence on predicted survival at the top, least influence on predicted survival at the bottom). The SHAP values are shown on the x-axis. SHAP values quantify the contribution of each predictor to the model estimate of mortality risk, in deviation from the mean prediction. The grey vertical line represents a zero-impact mean prediction. Positive values (to the right of the zero-impact line) are associated with increased mortality risk and negative values (to the left of the zero-impact line) with reduced mortality risk. As the data points for each predictor move further from the vertical line, the greater the impact of this predictor on expected survival becomes. The colour spectrum (blue to red) across the SHAP value scatter shows the value of the predictor value, with blue indicating lower values and red signifying higher values.  This relationship is most easily visualised in the `age\_category` and `chess\_scale\_score`, where higher values (older age or worse health status respectively) are both red and associated with high positive SHAP values (large impact on the prediction of increased mortality risk).

We also gain insights and witness the asymmetric effects that certain predictors and their values exert in influencing the final predicted risk scores. For instance, `rx\_risk\_score`, `poor\_eating\_or\_lack\_of\_appetite` and the `pressure\_ulcer\_risk\_score`, exhibit a much stronger effect on elevating the predicted risk scores as their predictor values increase, while the reverse effect is smaller on reducing risk as their predictor values decrease. This is also in line with expectations, since for example, evidence of poor eating or a lack of appetite ought to have a greater effect on the model than a lack of evidence thereof. Other notable predictors are `specific\_health\_conditions`and `cognitive\_performance\_scale\_score` which are largely consistent in signalling that a deterioration (higher values) in these predictors tends to also increase the predicted risk. However, a less coherent signal accompanies the 'depression\_rating\_scale\_score', `skin\_integrity\_score`, `faecal\_incontinence`  and `falls\_history' (investigated further below in dependence plots) with some signs of ambivalence with respect to predicted risk scores as the underlying predictor values change.

\begin{figure}[hbt]
    \centering
    \begin{subfigure}[b]{0.34\textwidth}
        \includegraphics[width=\textwidth]{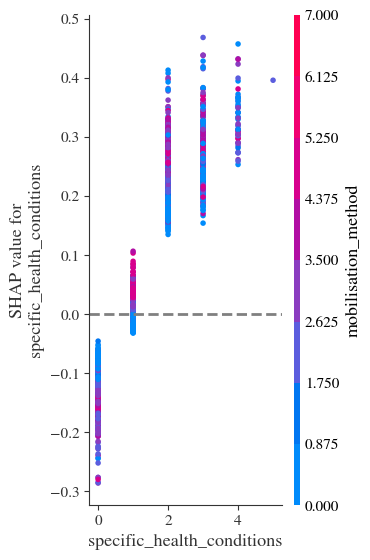}
        \caption{}
    \end{subfigure}%
    \begin{subfigure}[b]{0.34\textwidth}
        \includegraphics[width=\textwidth]{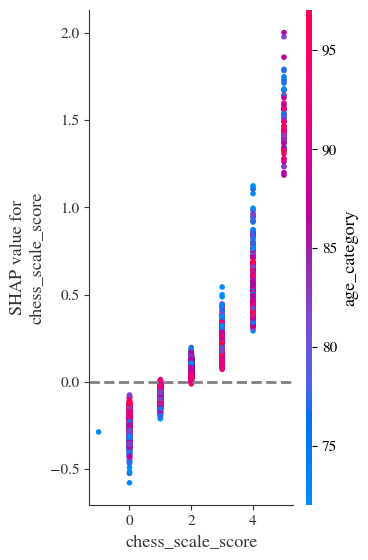}
        \caption{}
    \end{subfigure}%
    \begin{subfigure}[b]{0.34\textwidth}
        \includegraphics[width=\textwidth]{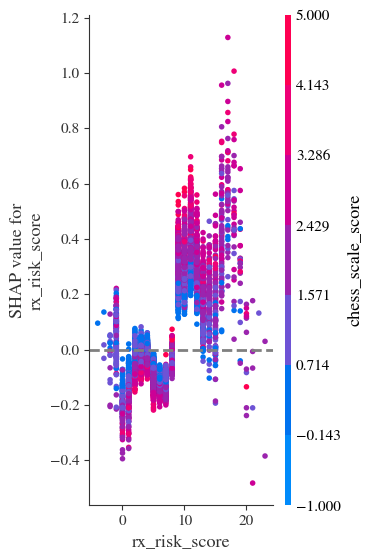}
        \caption{}
    \end{subfigure}%

    \begin{subfigure}[b]{0.34\textwidth}
        \includegraphics[width=\textwidth]{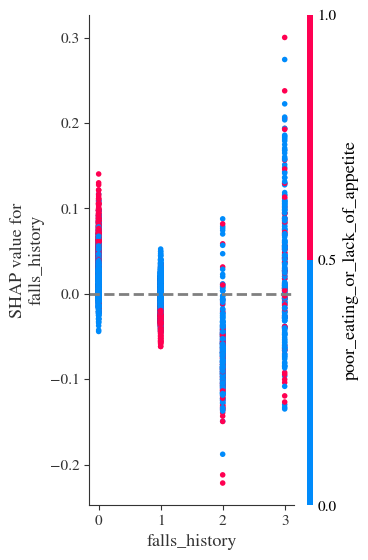}
        \caption{}
    \end{subfigure}%
    \begin{subfigure}[b]{0.34\textwidth}
        \includegraphics[width=\textwidth]{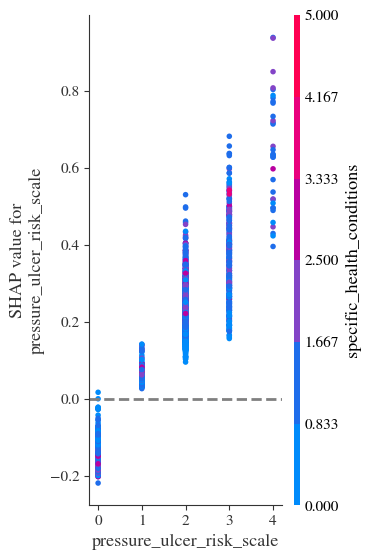}
        \caption{}
    \end{subfigure}%
    \begin{subfigure}[b]{0.34\textwidth}
        \includegraphics[width=\textwidth]{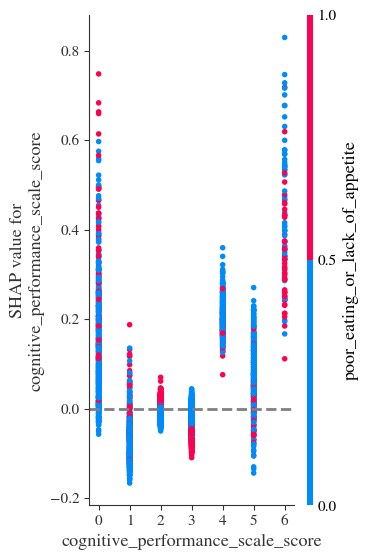}
        \caption{}
    \end{subfigure}%

    \caption{Dependence plots showing pairwise interactions between a selection of predictors and how their interactions affect patient risk predictions by the XGBoost model.}
    \label{fig:dependence}
\end{figure}

A set of six Dependence Plots are depicted in Figure \ref{fig:dependence}a-f, offering a deeper understanding of pairwise interactions between a selection of predictors. These visualizations depict how interactions of pairs of predictors influence the prediction of risk as their underlying values vary. As previously stated, these figures are based on the XGBoost model.
For each of the selected predictors, the SHAP tool automatically selects the most interactive corresponding predictor.

The x-axis represents the values for a chosen predictor, while the gradient colour bar on the y-axis represents the values of the counterpart predictor. The interaction of both is depicted with respect to the magnitude of the impact they exert on the final prediction. The dashed horizontal line represents a neutral effect on the model output. Points above this line indicate an increase in mortality risk, while the opposite holds for values below the dashed line. The relative distance from the dashed line indicates the magnitude of the effect exerted on the mortality risk.

We see in Figure \ref{fig:dependence}a that as the number of `specific\_health\_conditions` increases, the overall risk follows at an acute rate.  Despite high mobilization requirements in the absence of `specific\_health\_conditions`, there is no elevation in risk.  As soon as `specific\_health\_conditions` are observed, higher associated mobilisation requirements tend to also elevate the risk. With an increasing `chess\_scale\_score` in Figure \ref{fig:dependence}b, the mortality risk gradually increases. For lower values of the `chess\_scale\_score`, the interaction with increasing patient age tends to elevate overall risk. This relationship, however, does not seem to hold for higher values of the `chess\_scale\_score`. A distinct pattern emerges for the `rx\_risk\_score` predictor in Figure \ref{fig:dependence}c as its values increase. A score of less than 10 for `rx\_risk\_score` does not show a tendency to increase mortality risk. The interaction of lower values for this predictor with increasing values of frailty represented by the `chess\_scale\_score` tends not to elevate risk. However, an inflexion point occurs from 10 onwards for the `rx\_risk\_score`, at which point increasing values for both this predictor and the `chess\_scale\_score`, interact to significantly elevate the mortality risk. The patterns from the `falls\_history` predictor are less clear in Figure \ref{fig:dependence}d. The x-axis represents the history of falls categorized into four discrete values: '0' representing for 1 or no history of falls, '1' for 4 or less in the last 6 months, '2' for 5 or more in the last 6 months, and '3' for 3 or more falls in a one-month period. Risk is high for patients having 1 or no history of falls in cases where their `poor\_eating\_or\_lack\_of\_appetite` values are high. This relationship however does not hold as the frequency of the reported falls also increases. Meanwhile, the variance in the effect that the falls have on the risk disperses more greatly as falls increase, with an ambivalent relationship emerging with the `poor\_eating\_or\_lack\_of\_appetite` predictor. Figure \ref{fig:dependence}e, a clear and gradual rise in the risk can be observed for each increase in the `pressure\_ulcer\_risk\_score`, which is only slightly amplified with increasing values in the number of `specific\_health\_conditions`. Finally, Figure \ref{fig:dependence}f shows the interaction between the `cognitive\_performance\_scale\_score` and the `poor\_eating\_or\_lack\_of\_appetite` values. The `cognitive\_performance\_scale\_score` tends to have the highest effect on elevating risk for the lowest and highest scores of this predictor. Overall, high `poor\_eating\_or\_lack\_of\_appetite` values seem to have a larger interaction effect on increasing risk with lower values of `cognitive\_performance\_scale\_score`.

\subsubsection*{Clinical usage and application}

\begin{figure}[h!]
    \centering
    \includegraphics[width=0.74\textwidth]{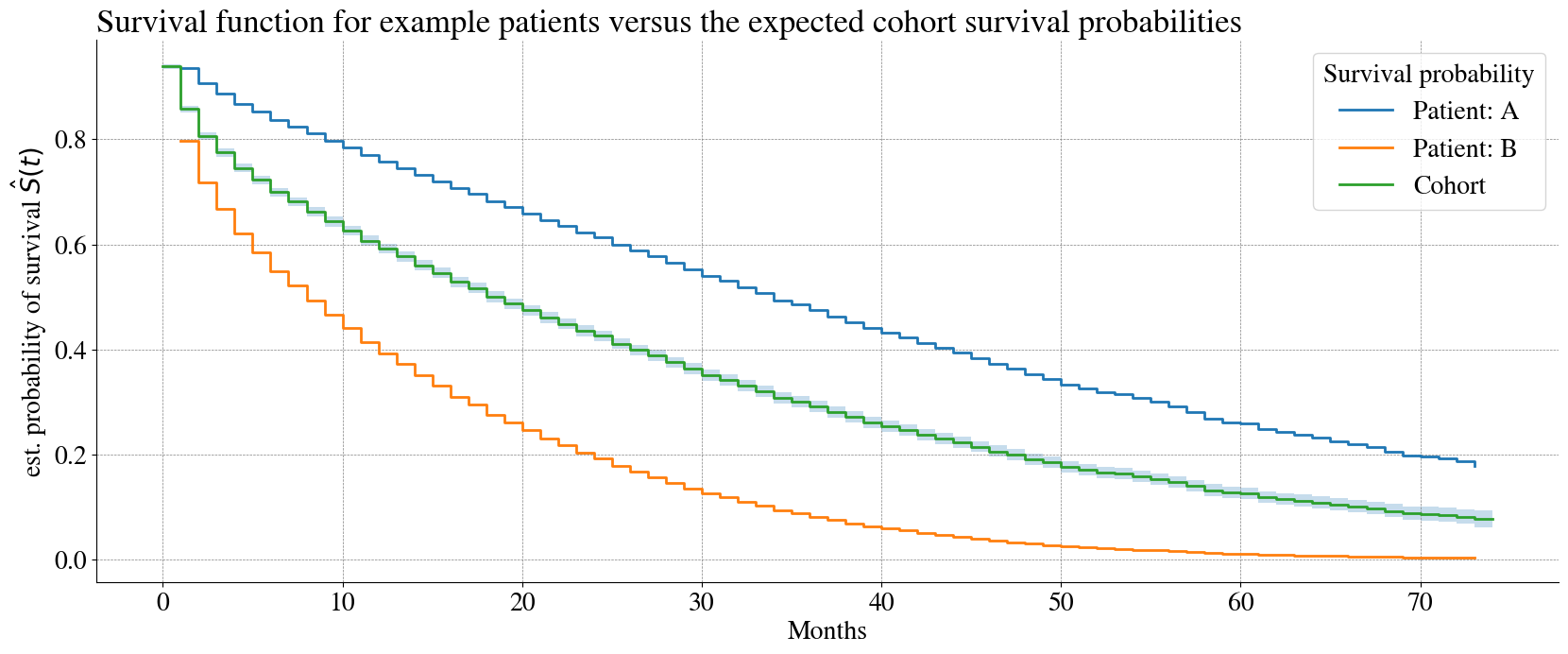}
    \caption{Patient survival function from the Gradient Boosting model depicting the risk of two exemplars versus the cohort.}
    \label{fig:example}
\end{figure}

The integration of survival analysis models into clinical practice is pivotal for informed medical decision-making. Here, we transition from theoretical modeling and inspection of a model from a high-level to a real-world application, using data from two anonymised patients as exemplars. The survival probability curve derived from the uncalibrated model and shown in Figure \ref{fig:example}, illustrates each patient's predicted survival trajectory in comparison to the cohort average. The figure indicates that the survival probabilities for patient B are significantly lower than those of patient A, and well below the cohort average across the entire timeframe of potential observation. This initial output allows clinicians to gauge individual patient risk in the context of broader population trends. However, while the initial outputs are useful, additional insights are needed to unpack how and why the model is arriving at different risk profiles for specific patients.

\begin{figure}[hbt]
    \centering
    \begin{subfigure}[b]{0.5\textwidth}
        \includegraphics[width=\textwidth]{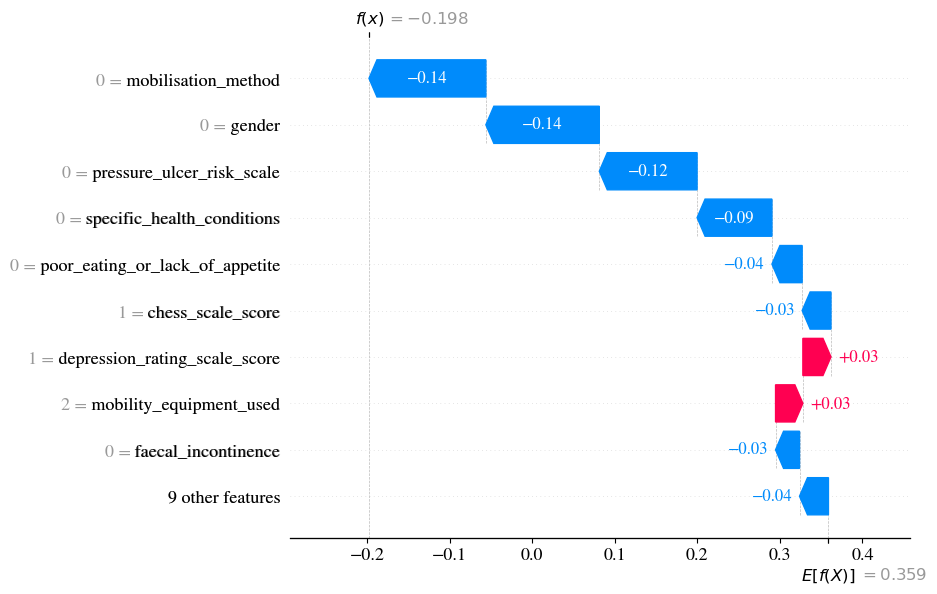}
        \caption{Patient A}
    \end{subfigure}%
    \begin{subfigure}[b]{0.5\textwidth}
        \includegraphics[width=\textwidth]{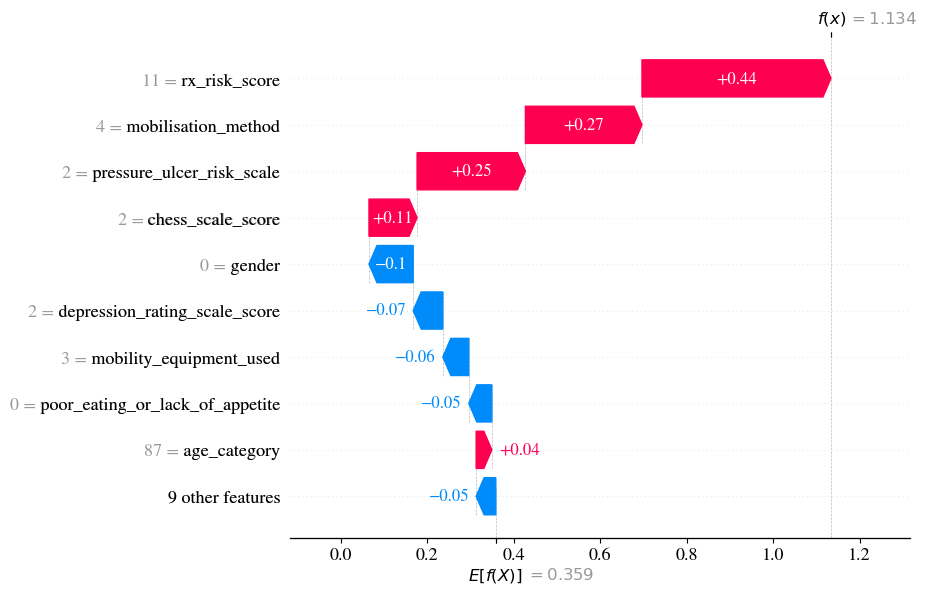}
        \caption{Patient B}
    \end{subfigure}%

    \caption{Waterfall plots from the XGBoost model showing interpretability for patient-level predictions using two hypothetical examples.}
    \label{fig:waterfall}
\end{figure}

Subsequently, the utilization of SHAP waterfall plots seen in Figure \ref{fig:waterfall}a-b offers patient-level model interpretability. These plots reveal what the key predictors and their values are and how they  influence the model's survival predictions for each patient. The emphasis here is on practicality: enabling clinicians to comprehend the underpinnings of the model's output, ensuring that its insights can be validated, trusted and ultimately integrated into tailored patient management strategies. Through this approach, we demonstrate the confluence of advanced analytical tools with clinical utility, underscoring their role in optimizing patient care. These plots are best interpreted from bottom-up. The y-axis shows the most impactful predictors with their values, and their relative contributions. The starting point on the x-axis is the average or the expected risk for the whole cohort. Each predictor pushes the risk to the left (to lower risk) or to the right (to increase risk) until all contributions are summed at the top row. For patient A in Figure \ref{fig:waterfall}a, we can see that the patient's result on the depression rating scale score and the use of mobility equipment increases their risk; however, their overall risk is significantly lowered by their independent mobilisation, low pressure ulcer risk score, low number of ongoing health conditions and their female gender. Meanwhile, for patient B in Figure \ref{fig:waterfall}b, it can be observed that their high risk is predominately driven by their high prescription risk score, limited mobility as well as their pressure ulcer risk score.

\subsection{Calibrated Time-specific Survival Models}

\begin{figure}[h!]
    \centering
    \includegraphics[width=0.6\textwidth]{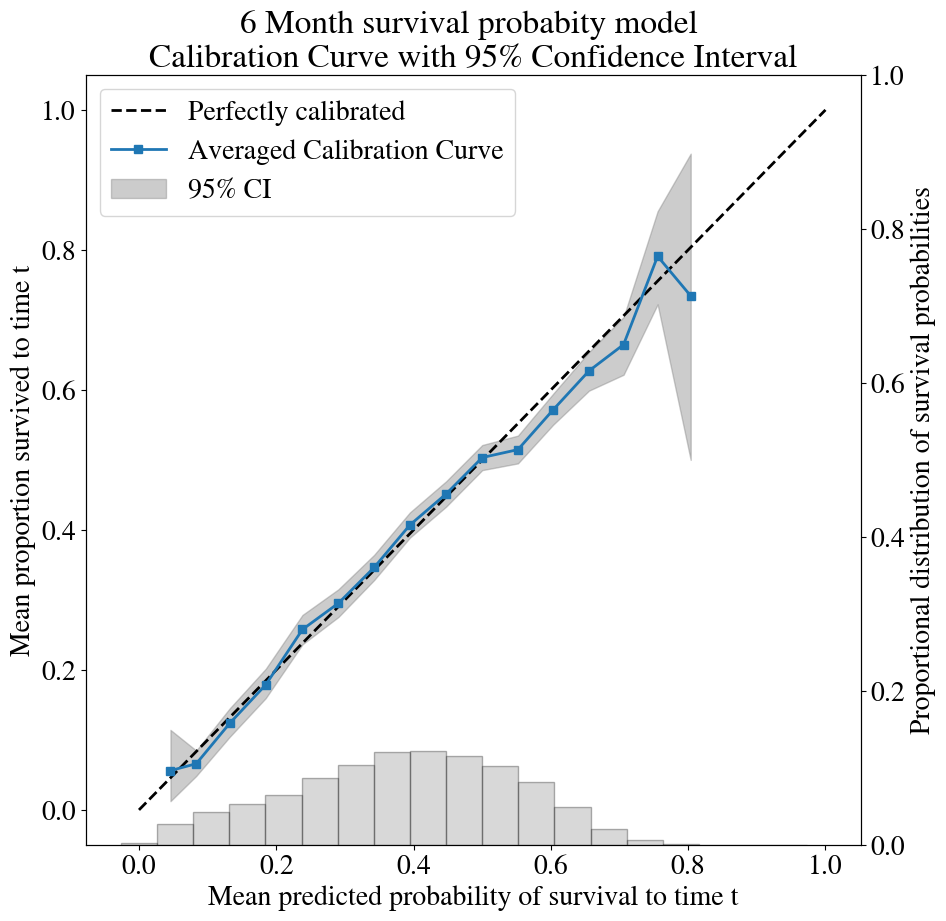}
    \caption{6-month Gradient Boosting model calibration plot}
    \label{fig:calibration}
\end{figure}

In an evaluation of survival models tailored for different forecast horizons, the calibration plots with 95\% confidence intervals for the target 6-month period (seen in Figure \ref{fig:calibration}) yields insights into the model's predictive accuracy. The plot features a calibration curve approximating the ideal 45-degree line, a sign of near-optimal calibration, where overall, the depicted model exhibits effective calibration from low to mid-range probabilities, while increased uncertainties around higher predicted survival probabilities (sub 0.8 probability) can also be seen. Beneath the calibration curve lies a histogram depicting the distribution of the predicted probabilities. The histogram intimates the density of predictions at different probability ranges. The shape of the distribution resembles a normal distribution with an albeit more pronounced tail for the lower probabilities, and a mean centred $\sim$0.4. From this, it can be assumed that the target model is effectively calibrated.

The performance metrics presented in Table \ref{tab:performance-metrics} offer a multi-perspective evaluation of Gradient Boosting models tailored for survival analysis across varying temporal horizons with respect to a range of metrics. For completion and comparisons, the accuracies of 1-, 3- 6- and 12-month calibrated models are shown. The Dynamic AUROC serves as an emblematic metric for assessing a model's discriminative capability. The observed downward trend in AUROC values as we move from short-term to longer-term forecasts is indicative of an interplay between model sensitivity and the inherent heterogeneity of patient trajectories over time. A declining AUROC is often attributed to the increased stochasticity of long-term forecasts, as can be seen in the table when contrasting the 1- and 12-month forecast accuracies.

\begin{table}[hbt]
\centering
\caption{Performance metrics of calibrated Gradient Boosting models for time-specific forecasts.}
\label{tab:performance-metrics}
\begin{tabularx}{\textwidth}{rllll}
\toprule
Forecast & Dynamic AUROC (95\% CI) & IBS (95\% CI) & C-index (95\% CI) & Harrell (95\% CI)\\
\midrule
\text{1-month} & 0.794 (0.789-0.799) & 0.296 (0.294-0.299) & 0.715 (0.712-0.717) & 0.674 (0.672-0.676) \\
\text{3-month} & 0.765 (0.762-0.768) & 0.280 (0.280-0.281) & 0.717 (0.716-0.719) & 0.676 (0.674-0.678) \\
\text{6-month} & 0.746 (0.744-0.749) & 0.259 (0.258-0.261) & 0.716 (0.714-0.718) & 0.675 (0.673-0.677) \\
\text{12-month} & 0.726 (0.723-0.729) & 0.239 (0.238-0.241) & 0.720 (0.718-0.722) & 0.680 (0.677-0.682) \\

\bottomrule
\end{tabularx}
\label{tab:timedependentresults}
\end{table}

The IBS serves as a gauge for model calibration. Though the observed trend of declining AUROC values over increasing prediction horizons aligns with the prevailing literature, signaling a diminishing discriminative power for long-term forecasts,  intriguingly the model’s IBS values improve (decrease) concurrently. This is  counterintuitive given the conventional wisdom that long-term forecasts usually suffer from poor calibration. This paradox can however be explained through the lens of the bias-variance tradeoff: the results suggest that as the models become better calibrated over time, their variance reduces, thereby increasing bias and consequently reducing the models' discriminative power.

The C-index and Harrell's index are used for their robustness in quantifying a model's ability to correctly rank-order individual risks. The results indicate the models' stability with respect to this across all forecast horizons. This is in contrast to the Dynamic AUROC, which demonstrates a deterioration as the prediction window extends. The stability in the C-index and Harrell's C-index could be indicative of the model's preserved efficacy in ranking the relative risk between individuals over time, even as its ability to separate the classes of events and non-events diminishes (evidenced by the declining AUROC). This points to an interplay between how the model assigns ordinal ranks to individual survival probabilities versus its performance in the classification of events. Thus, the observed stability in C-index and Harrell's C-index adds a layer of confidence in the model's utility for tasks that require risk stratification over dichotomous classification, a distinction that has important implications in a clinical setting.

The results of the calibrated Gradient Boosting model exhibits attributes that are tied to the temporal granularity of its prognostic estimates. Although the Dynamic AUROC, a traditional metric of discriminative power, reveals a time-sensitive attenuation, this need not be misconstrued as a universal decline in model efficacy. Importantly, the model's calibration, captured through the IBS, and its discriminatory consistency, as evidenced by stable C-index and Harrell's C-index metrics remain largely unaltered across varying forecast horizons. This observed dichotomy between discriminative power and risk-ranking capacity necessitates a departure from monolithic evaluation frameworks. It underscores the imperative for a multi-metric paradigm that captures the multi-dimensional attributes of survival models.

\subsection{Clinical Validity}

\begin{figure}[hbt]
    \centering
    \includegraphics[width=0.74\textwidth]{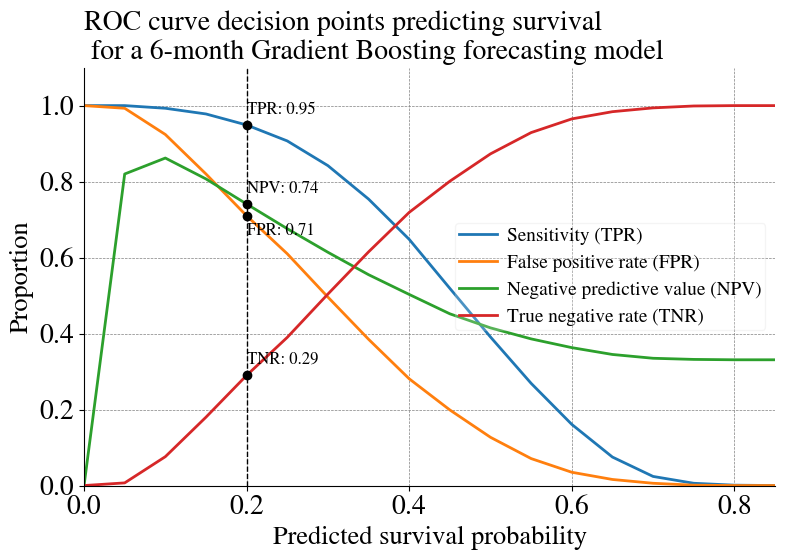}
    \caption{ROC curve of the calibrated Gradient Boosting 6-month model predicting survival probabilities. An example operational threshold of 0.2 for predicting patient survival is highlighted as a practically useful decision point from a clinical perspective.}
    \label{fig:roc6m}
\end{figure}

The ROC curve illustrated in Figure \ref{fig:roc6m} for the Gradient Boosting model offers an empirical framework for clinicians to discern patients at a heightened risk of mortality within 6 months post-admission to long-term care facilities. It illustrates the balance between specificity (true negative rate, TNR) and sensitivity (true positive rate, TPR) achieved by varying the prediction threshold. The figure highlights the clinical implications of adopting a 0.2 survival probability threshold, equating to an 80\% risk of death within the specified period, thereby guiding interventions. Performance metrics presented are derived from validation on a separate dataset, ensuring a robust appraisal of the model's predictive capabilities.

\begin{itemize}
  \item \textbf{Sensitivity / True Positive Rate (TPR)}: This metric quantifies the model's ability to identify actual survivors, with a threshold of 0.2 yielding a 95\% TPR. This means that 95\% of patients who survive beyond six months are accurately predicted by the model.
  \item \textbf{Negative Predictive Value (NPV)}: NPV assesses the accuracy of the model in predicting non-survival. At a threshold of 0.2, the NPV is 74\%, indicating that among those predicted not to survive, 74\% did not survive past six months.
  \item \textbf{False Positive Rate (FPR)}: FPR reflects the proportion of non-survivors incorrectly predicted as survivors. With a FPR of 71\%, the model erroneously predicts survival in 71\% of cases where the patient does not survive six months.
  \item \textbf{Specificity/True Negative Rate (TNR)}: This metric measures the model's precision in identifying non-survivors.  Although a specificity of 0.29 is low, this level of caution is appropriate in a situation where under-estimating the likelihood of death within six months carries potentially less clinical and ethical risk than an over-estimation of risk.
\end{itemize}

\section{Discussion}
We are not the first investigators to develop a prognostic model for people admitted to residential aged care. The work we present here builds on and extends the work of many others\cite{ogarek2018minimum, porock2010mds, flacker2003mortality, mitchell2010prediction, mitchell2004estimating, niznik2018adaptation, bicknell2020study}.  Our contribution has been to use advanced machine learning algorithms and eXplainable AI on a very large set of standardised health data to generate a clinically useful decision support tool.

We have demonstrated the feasibility of developing a survival model based on data acquired at the time of admission to residential care. People admitted to residential care have complex medical and nursing requirements and initial clinical assessments can take some time to complete, so we included data acquired up to 1 month post-admission. Our cohort included people admitted over a six-year period and the digital health record evolved and expanded during this time, with new assessments added to meet new regulatory and clinical requirements.
 Consequently, we struck the issue of substantial amounts of missing data for potentially useful variables we might have wished to include, a common problem in "real life" data sets, as opposed to data prospectively collected for research purposes. The predictors in our final model thus reflect a pragmatic balance struck between including the most consistently recorded items and the most relevant clinical details, and our desire to retain the maximal amount of training data.

We investigated the efficacy of a variety of traditional and machine learning approaches by conducting rigorous repeated experiments with seven different algorithms, using test-training splits to minimise over-fitting, having evaluated all models using appropriate performance metrics. We found minimal differences between the top-performing models. The three machine learning ensemble models (GB, RF and XGB) consistently outperformed other algorithms on most evaluations. CoxPH outperformed other traditional statistical methods for generating survival curves, but was not as high-performing as the ensemble models.  The evaluation metrics for these models confirm satisfactory discriminatory power and robust statistical stability at a level of accuracy that matches or exceeds other prognostic models developed in residential aged-care populations \cite{kruse2010using, van2007prediction, zhang2023prediction}.

We used an uncalibrated XGB model to generate a continuous function showing survival probability at every time point up to six years post-admission.  This type of survival curve, generated for the entire cohort,  for selected categories within the cohort (by age, gender,  or CHESS score for example), or for individual patients, is familiar to most clinicians.  Overlaying a survival curve for an individual resident with the survival curve for their cohort of peers creates an instantly usable visual aid that could be used to inform discussions about prognosis with a patient or their family, while avoiding being overly definitive or confronting.

We then calibrated the top-performing model (GB) to generate  a point-estimate of survival probability at six months post-admission and integrated a closely related uncalibrated variant model (XGB) with SHAP tools to produce visualizations that illuminate the internal decision-making processes within the model and the complex interactions between predictors.  We calibrated to six months as it is a prognostic time frame where open discussions about patient preferences for end-of-life care become appropriate and necessary, as reflected in the recommendation to use the InterRAI palliative care assessment for people in residential care with a life expectancy of six months or less \cite{interrai2023palliative}. Funding for palliative services in a primary care setting is only provided for people expected to die within six months. This funding is rarely accessed by people admitted to residential aged care due to the frequent absence of any specific terminal diagnosis.  More accurate prognostication in these individuals has the potential to address an equity issue by improving their access to appropriately funded palliative care services.

There are other strong arguments for providing an accurate prognosis for older people admitted to residential care. Health professionals working in aged care are often reluctant to discuss prognosis with residents and their families for a variety of reasons despite the fact that more than one-third of people admitted for long-term care die within six months of their arrival. Many of these individuals are poorly served by our failure to openly acknowledge their limited life expectancy and are subjected to treatments that neither extend nor enhance the quality of their remaining life. They and their families are often inadequately prepared for death, resulting in a traumatic terminal experience for the resident and complex grief in the survivors.

In an ideal world, sensitive but realistic conversations about prognosis and expected goals of care would occur with every person and their family as a routine part of their admission to aged care. In this study, we have applied the advanced analytical techniques offered by machine learning and XAI to create useful visual aids that could support these conversations.   Ultimately, however,  the value of any tool is only realised in its application.  Decision support tools increase patient autonomy and enhance clarity in healthcare discussions,  but only if healthcare providers choose to use them.

\subsection{Study limitations}

This study has several limitations that we acknowledge.  Firstly, the dataset contained significant missing values for many variables. While multiple imputation methods like MICE were utilized to address this, the relatively high proportion of missing data for some predictors means uncertainty persists regarding their true values and subsequent impact on the models. Specifically, this likely contributed to the ambiguous signals seen for certain variables in the SHAP analyses, hampering model interpretability.

Secondly, the study relied solely on data collected around the time of admission to the aged care facilities, meaning the prognostic models are optimal for short-term predictions during this period but become less reliable for longer-range forecasts.  Limiting data collection to a single time point also precluded incorporating temporally dynamic variables, such as changes in functional capacity, mobility, falls frequency, and appetite, that are likely to influence mortality risk trajectories beyond the initial admission window.

Thirdly, the cohort consisted of a heterogeneous mix of patient populations from a single private aged care provider. Thus, the model may exhibit limited generalizability to other provider settings that diverge in their patient demographics and data collection protocols. Evaluating model transportability across diverse validation datasets would further confirm its broader utility.

Finally, the study establishes technical efficacy but lacks an assessment of real-world clinical implementation factors. Pragmatic clinical trials are imperative to evaluate the proposed models' perceived utility among staff end-users and tangible impacts on workflows and decision-making prior to actual deployment in day-to-day clinical practice. User-centred design principles could help optimize the integration and presentation of model insights at the point of care.

\subsection{Future work}

Multiple promising avenues exist to build upon this work and to enhance the predictive capabilities and clinical utility of data-driven prognostic models in the aged care setting. Firstly, incorporating dynamic time-varying covariates into risk forecasts could improve accuracy, especially for predictions taking place after the initial admission time point. Expanding the feature space with descriptive variables capturing changes in patients' underlying conditions over time may also add valuable revisions to patient risk trajectories. Additionally, testing transportability across diverse datasets is valuable to confirm the generalizability of the models. Conducting clinical trials would provide useful real-world validation of the models' acceptability, trustworthiness, and measurable impacts on healthcare providers and patients.

Secondly, exploring emerging data modalities and advanced techniques offers opportunities to further enhance prognostic capabilities. Incorporating genomic biomarkers or medical imaging data could identify new prognostic subtypes amenable to personalized interventions. Applying natural language processing to intake records, clinician notes or patient narratives may also uncover novel insights not captured through structured variables. As such innovations are pursued, developing standardized guidelines will be imperative to ensure responsible development, evaluation, and monitoring that safeguards against potential pitfalls and harms. While this study establishes a methodological foundation on the use of machine learning and interpretable AI tools, important work remains in translating prognostic modelling advances into safe, effective, and patient-centred clinical decision support tools that measurably improve end-of-life care delivery.

\section{Conclusion}

This study demonstrates the feasibility of developing robust, predictive survival models for older individuals admitted to residential care. By leveraging advanced machine learning algorithms and incorporating explainable AI (XAI) techniques, the study not only achieved accurate survival predictions but also ensured transparency and interpretability in its models and their outputs. The integration of a wide range of variables collected at the point of admission to long-term care reflects the complex nature of health decline in the very old, extending survival analysis beyond traditional predictors. Our findings reveal that machine learning models coupled with XAI, offer a significant advantage in clinical decision-making by balancing high predictive accuracy with clear interpretability. The use of SHAP waterfall plots for individual-level risk assessment and calibration of predictions to specific time horizons, such as six months, enables healthcare professionals to make informed, personalized decisions for their patients. The proposed predictive framework and tools represent a step forward, offering a comprehensive approach that combines predictive accuracy with an understanding of the underlying factors influencing patient survival. This approach not only supports clinical decision-making around appropriately targeted palliative care but also enhances trust in AI-driven prognostic tools in healthcare.

\bibliographystyle{unsrtnat}


\end{document}